\documentclass[10pt,twocolumn,letterpaper]{article}
\usepackage{iccv}
\usepackage{times}
\usepackage{epsfig}
\usepackage{graphicx}
\usepackage{amsmath}
\usepackage{amssymb}
\usepackage{subfig}
\usepackage{enumitem}
\DeclareGraphicsExtensions{.png,.jpeg,.pdf}

\iccvfinalcopy
\def\iccvPaperID{****} 

\begin{document}

\title{Neural Style Transfer for Point Clouds}

\author{Xu Cao$^{\star }$ \qquad Weimin Wang$^{\dagger}$ \qquad Katashi Nagao$^{\star}$ \\
$^{\star}$ Graduate School of Informatics, Nagoya University, Japan \\
    $^{\dagger}$ National Institute of Advanced Industrial Science and Technology (AIST), Japan\\
{\tt\small sou@nagao.nuie.nagoya-u.ac.jp, weimin.wang@aist.go.jp, nagao@nuie.nagoya-u.ac.jp}
}

\maketitle

\begin{abstract}
How can we edit or transform the geometric or color property of a point cloud? In this study, we propose a neural style transfer method for point clouds which allows us to transfer the style of geometry or color from one point cloud either independently or simultaneously to another. This transfer is achieved by manipulating the content representations and Gram-based style representations extracted from a pre-trained PointNet-based classification network for colored point clouds. As Gram-based style representation is invariant to the number or the order of points, the same method can be extended to transfer the style extracted from an image to the color expression of a point cloud by merely treating the image as a set of pixels. Experimental results demonstrate the capability of the proposed method for transferring style from either an image or a point cloud to another point cloud of a single object or even an indoor scene.
\end{abstract}

\section{Introduction}
\label{sec:intro}
Since 3D point clouds are usually the raw data that most 3D sensors directly acquire, it is an important 3D representation in various applications like robotics, autonomous driving and geographical information processing. A point cloud not only reveals the geometry information of 3D objects, but also it reflects the color information of the 3D objects. As LiDAR sensors and RGB-D cameras become more and more prevailing, it has become easier to obtain point cloud contents nowadays. 
 
To make point clouds more suitable for downstream applications, several point cloud processing tasks have been studied including denoising  \cite{pcdDenoising}, registration \cite{pcdRegistration}, to name a few. However, there are few studies on the point cloud editing, i.e., modify the geometry or color property of a point cloud. This gap may be caused by the intrinsic difficulty of point cloud editing as hundreds of thousands of points are laying in 3D space, causing difficult human-computer interaction. 
 
In this study, we propose a neural network based style transfer method for point clouds which allows us to modify the geometry or the color property of a point cloud by providing a target style point cloud or image. A PointNet \cite{pointnet} based classification network for colored point clouds is designed to extract feature representations of the geometry and color property independently. By matching the feature representation of the geometry or the color from a style point cloud, the geometry, the color or both of a content point cloud can be stylised. This method allows us to edit a point cloud just by providing another. Moreover, the same method can be extended to transferring the style of an image onto the color expression of a point cloud by treating the image as a set of pixels.
To summarize, our contributions are shown as follows:
\begin{enumerate}
\item We perform style transfer directly onto point clouds by matching feature representations extracted from a PointNet based network.
\item We demonstrate that our method can transfer the style of either the geometry or color property between single object point clouds, which may comprise a different number of points.
\item We extend the same method to stylise the color property of point clouds from images.
\end{enumerate}

\section{Related work}
\label{sec:related work}
\textbf{Neural style transfer between images}  Neural style transfer aims at seeking a stylised image that preserves the content of a content image with its style from a style image by utilizing feature representations extracted from neural networks. Gatys et al. \cite{gatys2016image} propose a style transfer method by iteratively updating an image such that its content representation and style representation extracted from VGG \cite{simonyan2014vgg} simultaneously match with that of a content image and a style image. This online optimization algorithm suffers from efficiency issue. To address the issue, a bunch of follow-up work additionally trains a neural network which can generate the stylised image with only one feed-forward pass. Depending on how many types of styles can one network generate, this type of method can be classified into three subclasses: per-style-per-model \cite{johnson2016fastNST,ulyanov2016fastNST}, multiple-style-per-model \cite{li2017multistyle,zhang2017multi} and arbitrary-style-per-model \cite{arbitraryStyle}. 

Gatys et al. \cite{gatys2016image} utilize the Gram matrix of feature maps to represent the style of an image, and this almost becomes an unsuspicious standard. Li et al. \cite{li2017MMD} point out the matching of Gram-based style representation is equivalent to minimizing Maximum Mean Discrepancy (MMD) between feature maps of style images and stylised images, and demonstrate other distribution alignment options for style transfer. 

Our method follows the scheme in \cite{gatys2016image}; however, unlike the case of image style transfer where pixels are laying in fixed grids, and only RGB color can be modified, both the geometry and color property of point clouds can be stylised by our method.

\noindent
\textbf{Point set representation learning} Learning good representations directly from point clouds in an end-to-end fashion is a challenging task due to its irregular data structure. PointNet \cite{pointnet} is a pioneering work in addressing this problem. To address the variant order issue of an point cloud, PointNet first applies a shared multi-layer perceptron (MLP) to expand each 3D point to higher dimensional feature space, and then use a symmetric function, e.g., max-pooling, to aggregate information along feature axis. This aggregation results in a global feature representing the overall shape of the point cloud. PointNet++ \cite{PointNet++} further improved PointNet by recursively applying PointNet to points in local regions to capture local structures at different scales. Similarly, PointCNN \cite{li2018pointcnn} designs an  $X$-conv operation to overcome irregularity, and PointSIFT \cite{jiang2018pointsift} designs an orientation encoding unit to extract multi-scale representations from point clouds. SpiderCNN \cite{spiderCNN} encodes the spatial difference between a point and its neighbours into filter functions and extend transitional convolutional operation to irregular point sets.

In this work, we modify PointNet \cite{pointnet} to extract feature representations for colored point clouds and demonstrate how can we utilize the feature representations to perform point cloud style transfer.

\section{Method}
\label{sec:method}
In this section, we first describe the architecture of the model for colored point cloud classification; then we explain how to utilize the feature representations extracted from the network for style transfer between point clouds. Finally, we explain how to extend the same method to transfer the style of images onto the color expression of point clouds.
\subsection{Classification network for colored point clouds}
\label{sec:classification_network}
We adopt a PointNet \cite{pointnet} based neural network to extract feature representations for point clouds. To extract a global feature for a point cloud that is invariant to the order of points, PointNet applies shared MLP to all points to extend their dimensions and utilizes a symmetric function (max pooling) to aggregate a global feature vector. 

\begin{figure}[htb]
\begin{center}
\subfloat[][]{\includegraphics[width=\linewidth]{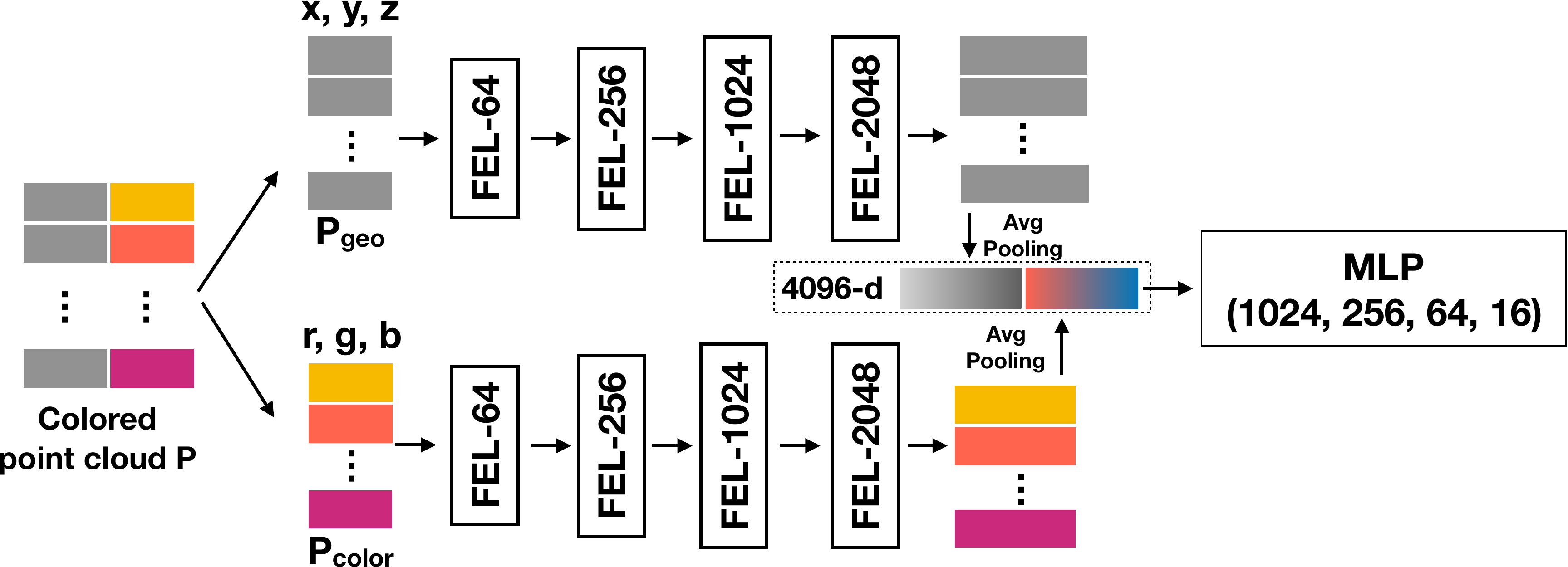}\label{model}}\\
\subfloat[][]{\includegraphics[width=\linewidth]{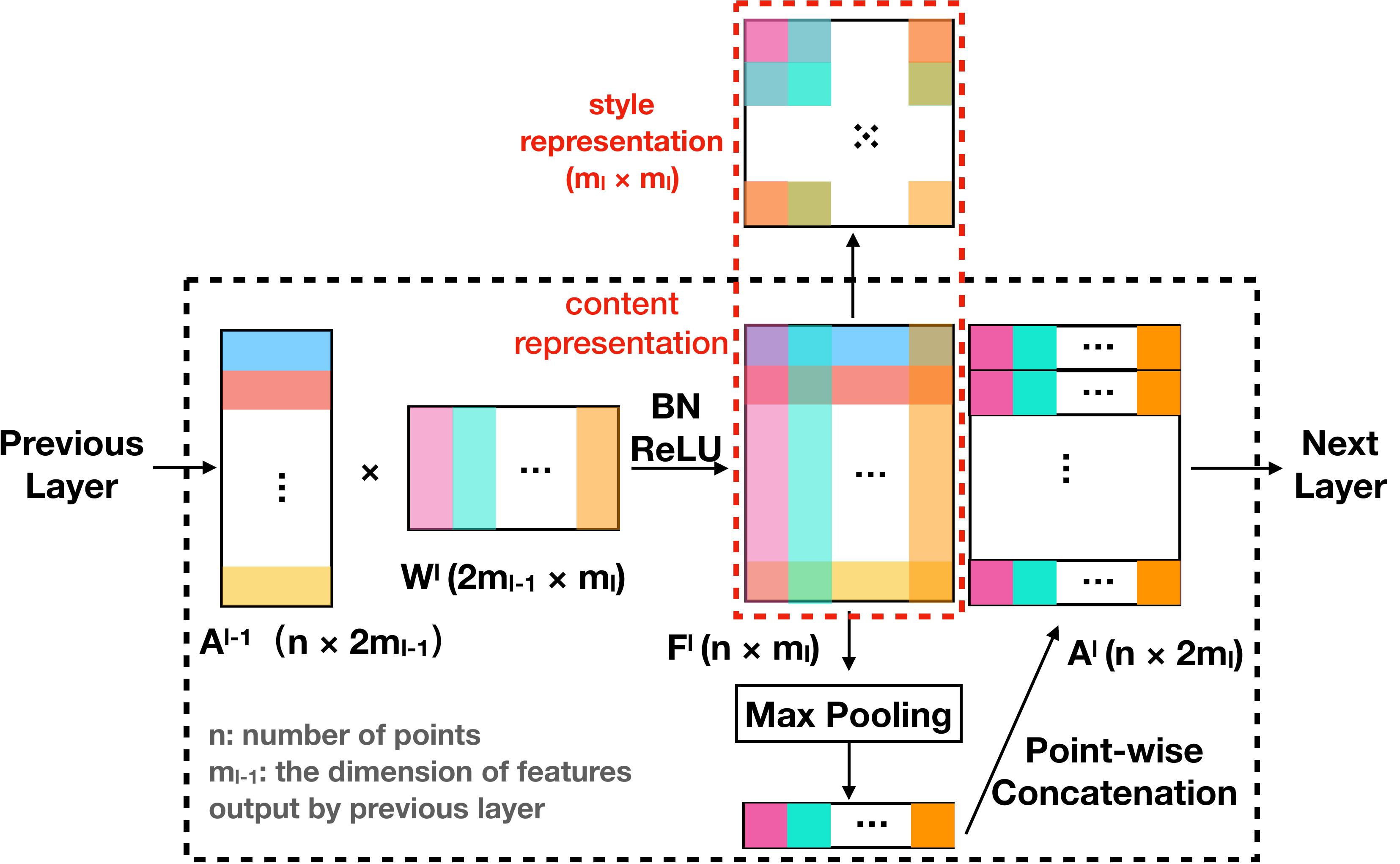}\label{FEL}}
\end{center}
\caption{Network architecture. \protect\subref{model} The geometry and color property of a point cloud are separately processed by a stack of FELs; then two global features are extracted and concatenated into one 4096-d vector, which was further processed by a MLP.  \protect\subref{FEL} Feature Encoding Layer (FEL). The numbers within brackets denote the dimensions of matrices. We first compute $F^l$ by multiplying each row of the input $A^{l-1}$ with the weight matrix $W^l$. Then, the maximum vector of $F^l$ is calculated and concatenated to each row of $F^l$. This concatenation results in $A^l$, the output of this layer and the input to the next layer. $F^l$ is taken as content representation while the Gram matrix of $F^l$ is taken as style representation.}
\label{fig:model}
\end{figure}

We make two modifications to PointNet for colored point clouds classification. First, instead of applying one shared MLP, we apply two separate MLPs to the geometry and color property of the point cloud, respectively. Without loss of generality, we denote a point cloud $P$ as a $N \times 6$ matrix, in which each row is its XYZ coordinates and RGB colors, and the identity of P is invariant to the order of its rows.  We split $P$ into $P_{geo}$ and $P_{color}$, two $N \times 3$ matrices containing only XYZ coordinates and RGB colors, respectively. Then two global feature vectors independently extracted for $P_{geo}$ and $P_{color}$ are concatenated into one 4096-d vector, as shown in Fig.\ref{fig:model}(a). 

Second, as shown in the black dashed box in Fig.\ref{fig:model}(b), for the output of each shared MLP, we extract a global feature vector and concatenate it to each feature. This operation results in features containing both its local information and global information. We term the modified layer as feature encoding layer (FEL). The number after FEL in Fig.\ref{fig:model}(a) denotes the dimension of each feature after being passed that FEL. Suppose $A^{l-1}$ is the input of the $l$-th FEL, then $A^l$ is obtained by $\text{FEL}(A^{l-1};W^l)$, and $A^{0}$ is $P_{geo}$ or $P_{color}$. In our implementation, $l$ is up to 4.
\subsection{Style transfer between point clouds}
\label{sec:transfer_between_point}
To perform style transfer between point clouds, we follow the idea of neural style transfer \cite{gatys2016image}. Given a pre-trained feature extractor, a content point cloud $C \in \mathbb{R}^{{N_c} \times 6}$ and a style point cloud $S\in \mathbb{R}^{{N_s} \times 6}$, we try to seek a stylised point cloud $P\in \mathbb{R}^{{N_p} \times 6}$ that minimizes the following loss function:
\begin{equation}
\newcommand{\argmin}{\mathop{\rm arg~min}\limits}
\begin{split}
P_{geo}^{\mbox{*}}=\argmin_{P_{geo}} \alpha_{geo} L_{geo\_content} (P_{geo}, C_{geo})  \\+ \beta_{geo} L_{geo\_style}(P_{geo}, S_{geo})  
\end{split} 
\label{eq:geo_total}
\end{equation}
\begin{equation}
\newcommand{\argmin}{\mathop{\rm arg~min}\limits}
\begin{split}
P_{color}^{\mbox{*}}=\argmin_{P_{color}} \alpha_{color} L_{color\_content} (P_{color}, C_{color}) \\+ \beta_{color} L_{color\_style}(P_{color}, S_{color})  
\end{split}
\label{eq:color_total}
\end{equation}
, where $L_{\cdot\_content}$ measure the difference between the content representation of $P$ and $C$, and  $L_{\cdot\_style}$ compares the style representation of $P$ and $S$. $\alpha$ and $\beta$ balance the content component and style component in the stylised point cloud. 

Let $F^l (\cdot) $ denote the feature response directly after the activation function in the $l$-th FEL. The number of row in $F^l (\cdot) $ is decided by that of the input point cloud, and the dimension of each row is $m_l$. This feature response is considered as the content representation of the point cloud, as shown in the red dashed box in Fig.\ref{fig:model}(b).
We define the content loss for geometry and color independently as follows:
\begin{equation}
\begin{split}
L_{geo\_content}&(P_{geo}, C_{geo})=\\ 
&\sum_{l\in \{l_c\}}{||F^l(P_{geo})-F^l(C_{geo})||^2} 
\end{split}
\label{eq:geo_content}
\end{equation}
\begin{equation}
\begin{split}
L_{color\_content}&(P_{color}, C_{color})=\\
&\sum_{l\in \{l_c\}}{||F^l(P_{color})-F^l(C_{color})||^2} 
\end{split}
\label{eq:color_content}
\end{equation}
, where $\{l_c\}$ denotes the set of FEL layers from which we extract feature representations for computing content loss.

The style representation in $l$-th FEL is the Gram matrix of $F^l(\cdot)$, as shown in the red dashed box in Fig.\ref{fig:model}(b). Next, we explain how this works for feature representation extracted by FELs. Let $F^l_{:j}(\cdot)$ denote the $j$-th column in $F^l(\cdot)$. Each element in this column is the inner product between the $j$-th column in the weight matrix $W^l$ and its corresponding row in $F^l(\cdot)$. Thus, the $j$-th column in $W^l$ can be viewed as a filter, and $F^l_{:j}$ is the response to that filter. In other words, the $j$-th column in $W^{l}$ and  $F^l_{:j}$ is analogous to a convolution kernel and the feature map to that kernel in convolutional neural networks.  

Taking the analogy into consideration, we compute $G_{ij}(F^l(\cdot))$, the $ij$-th element of the Gram matrix for $F^l(\cdot)$ as the inner product of the $i$-th column and $j$-th column of $F^l(\cdot)$:
\begin{equation}
G_{ij}(F^l(\cdot)) = F^l_{:i} \cdot F^l_{:j} \label{eq: gram_point}
\end{equation}
We denote $G(F^l(\cdot)) \in \mathbb{R}^{m_l \times m_l}$as the whole Gram matrix, of which the dimension is solely decided by the number of columns in the weight matrix $W^{l}$. 

The style loss for geometry and color is defined independently as follows:
\begin{equation}
\begin{split}
L_{geo\_style}&(P_{geo}, S_{geo})=\\
&\sum_{l\in \{l_s\}}{||G(F^l(P_{geo}))-G(F^l(S_{geo}))||^2} 
\end{split}
\label{eq:geo_style}
\end{equation}
\begin{equation}
\begin{split}
L_{color\_style}&(P_{color}, S_{color})=\\
&\sum_{l\in \{l_s\}}{||G(F^l(P_{color}))-G(F^l(S_{color}))||^2} 
\end{split}
\label{eq:color_style}
\end{equation}, where $\{l_s\}$ denotes the set of FEL layers from which we extract feature representations to compute the style loss.

To seek $P_{geo}^{\mbox{*}}$ or $P_{color}^{\mbox{*}}$, we adopt a gradient-based optimization method, e.g., Adam optimizer \cite{Kingma2014AdamAM}, to iteratively update $P_{geo}$ or $P_{color}$ until its convergence.
\subsection{Style transfer from images onto point clouds}
\label{sec:style_transfer_image}
To transfer the style of an image to the color property of a content point cloud $C_{color}$, we merely treat an image as a set of pixel values $S_{color}$ and follow Eq.\ref{eq:color_total} to seek the stylised point cloud $P_{color}$.
This method works due to two advantageous properties of Gram matrix. First, the dimension of Gram matrix is solely decided by the number of columns in the weight matrix and irrelevant to the number of points. That is, the difference in the number of points in $S$ and $P$ doesn't disable the computation of Eq.\ref{eq:color_style}. Second, Gram matrix $G(F^l(S_{color}))$ is invariant to the order of rows in $S_{color}$. According to the two properties, by simply reshaping an image of dimension $H \times W \times 3$ into dimension $(H * W)\times3$, the same method discussed in Sec.\ref{sec:transfer_between_point} can be used to transfer the style of an image onto the color property of a point cloud.

Although this reshaping operation discards the spatial structure information of the image, experimental results show that the color distribution is captured and well transferred onto $C_{color}$.
\begin{figure}[t]
\begin{center}
\subfloat[][]{\includegraphics[width=\linewidth]{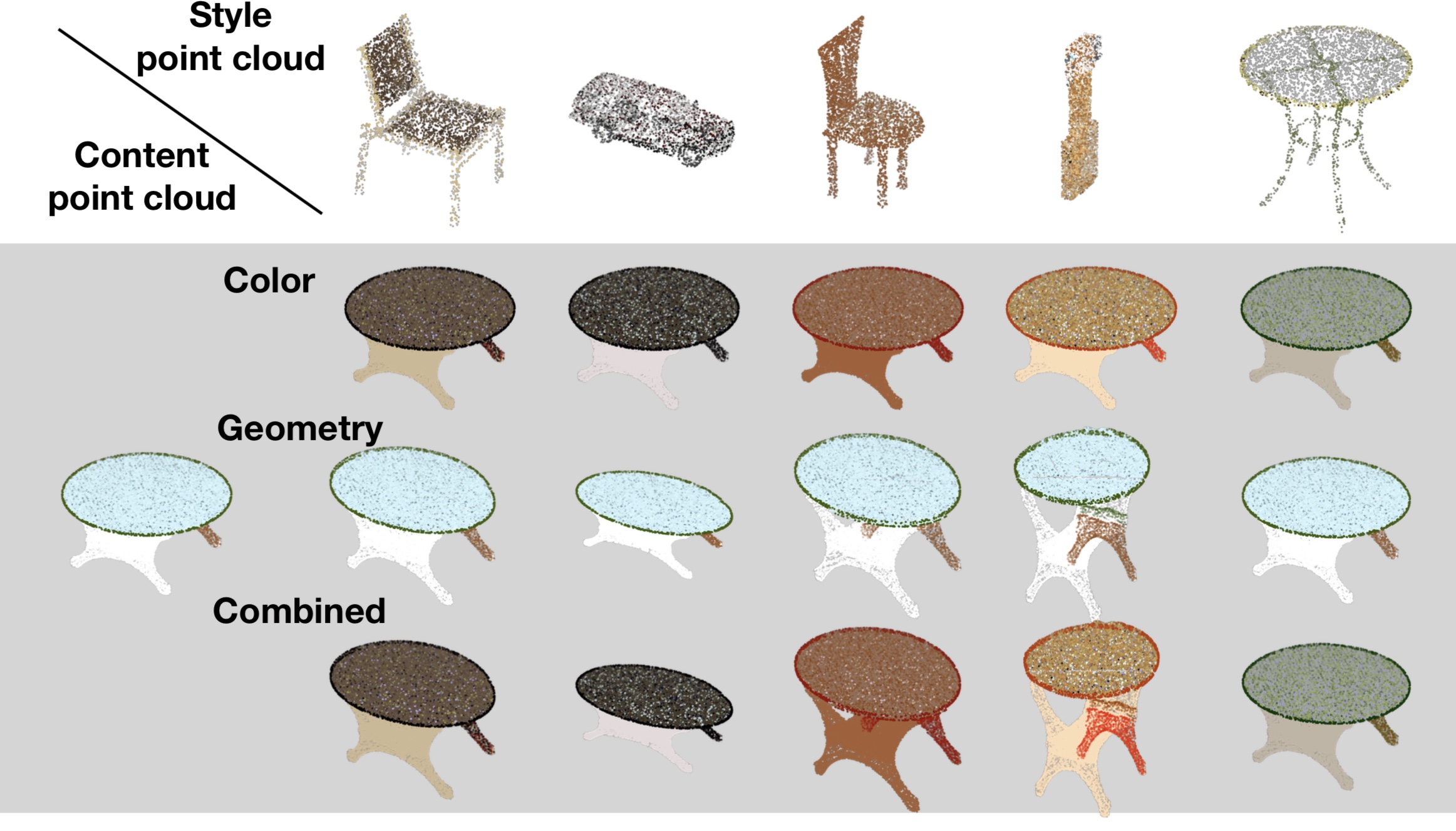}\label{either}}\\
\subfloat[][]{\includegraphics[width=\linewidth]{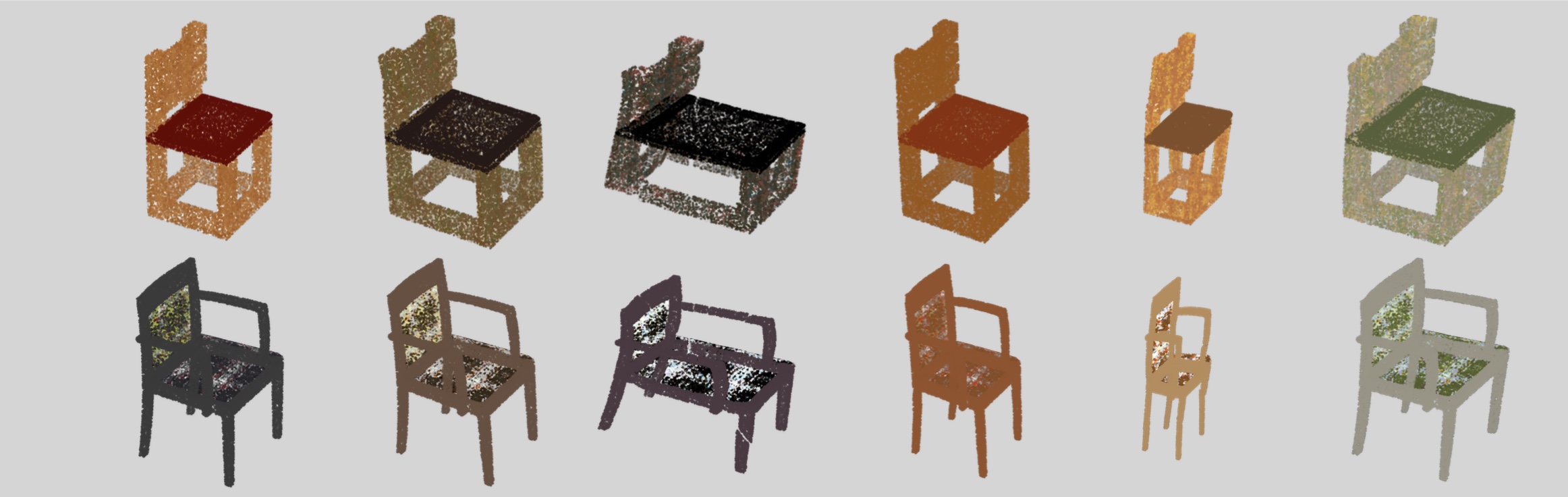}\label{both}}
\end{center}
\caption{Style transfer results between single object point clouds. \protect\subref{either} Either the geometry, the color property or both is stylised, and the points number of style point clouds is arbitrary. \protect\subref{both} More results where both the geometry and color property are stylised.}
\label{fig:pc_transfer}
\end{figure}
\section{Experiments}
\label{sec:experiments}
In Sec.\ref{sec:model_training}, we give training details of our proposed model for extracting representations of colored point clouds. In Sec.\ref{sec:qualitative_result}, we demonstrate qualitative results of the style transfer method. In Sec.\ref{sec:analysis_of_style_transfer}, we analyze different factors affecting style transfer results. In Sec.\ref{sec:ablation_study}, we perform ablation study on our model design. 
\subsection{Model setup for feature extraction}
\label{sec:model_training}
\subsubsection{Dataset}
To train a model that can encode both geometry and color of a point cloud, we use DensePoint dataset \cite{densepoint}. DensePoint is point cloud dataset built on top of ShapeNet \cite{shapenet2015} and ShapeNetPart Dataset \cite{shapenetpart}. There are 10454 colored point clouds of single objects across 16 categories; the dataset is split into 7612/1112/1730  training/validation/test subsets according to ShapeNet's setting.  
\subsubsection{Training settings}
DensePoint suffers from class imbalance problem. For example, there are over 3000 instances for table class while there are only less than 100 instances for several other classes. To remedy the problem, we use a combination of undersampling and oversampling techniques \cite{classimbalance}:  we undersample or oversample 320 instances for classes in which instances are over or under 320, respectively. For the value of XYZ coordinate and RGB colors in point clouds, we normalize them into the range between -1 and 1.

We train our model for 50 epochs with batch size 32. We use Adam optimizer \cite{Kingma2014AdamAM} with initial learning rate 0.01,  $\beta_1$ 0.9 and $\beta_2$ 0.999.  Batch normalization \cite{ioffe2015batch} is applied before activation functions in all layers except the last layer. Leaky ReLU with a fixed leakiness parameter 0.2 is used as activation functions. Dropout \cite{srivastava2014dropout} with keep ratio 0.7 is applied on last three fully-connected layers.
\subsection{Qualitative results of neural style transfer }
\label{sec:qualitative_result}
With the pre-trained network, we utilize it to extract content or style representations for point cloud style transfer. In this section, we demonstrate qualitative results of the proposed style transfer method.
\subsubsection{Style transfer between single object point clouds }
\label{sec:transfer_between_pcs}
Given a content point $C$ and a style point cloud $S$,  three factors affect the resulted point cloud $P$: the initialization strategy of $P$, the value of $\alpha$ and $\beta$ and the layers used for extracting representations. In this experiment, we initialize $P$ as $C$, let $\alpha_{geo} = \beta_{geo}= \alpha_{color} = 1, \beta_{color} = 100$, and extract content and style representations both from \{FEL-64\}. We adopt Adam optimizer with learning rate 0.01 to update $P$ and empirically set the update steps as 4000. The results are given in Fig.\ref{fig:pc_transfer}. In Fig.\ref{fig:pc_transfer}(a), we demonstrate two advantageous point of our method. First, the geometry or color property of a point cloud is independently stylised. Second, there is no constraint on the number of points in $C$ and $S$. In this experiment, $S$ contains 4096 points while $C$ contains 40k points. We give more qualitative results in Fig.\ref{fig:pc_transfer}(b). We find that in the case of geometry transfer, the overall shape style of $S$ is transferred; in the case of color transfer, the content color pattern is preserved while the overall color distribution is shifted towards that of style point clouds.
\subsubsection{Color transfer from images onto point clouds}
\label{sec:color_transfer_from_images}
\begin{figure}[tb]
\begin{center}
\subfloat[][]{\includegraphics[width=\linewidth]{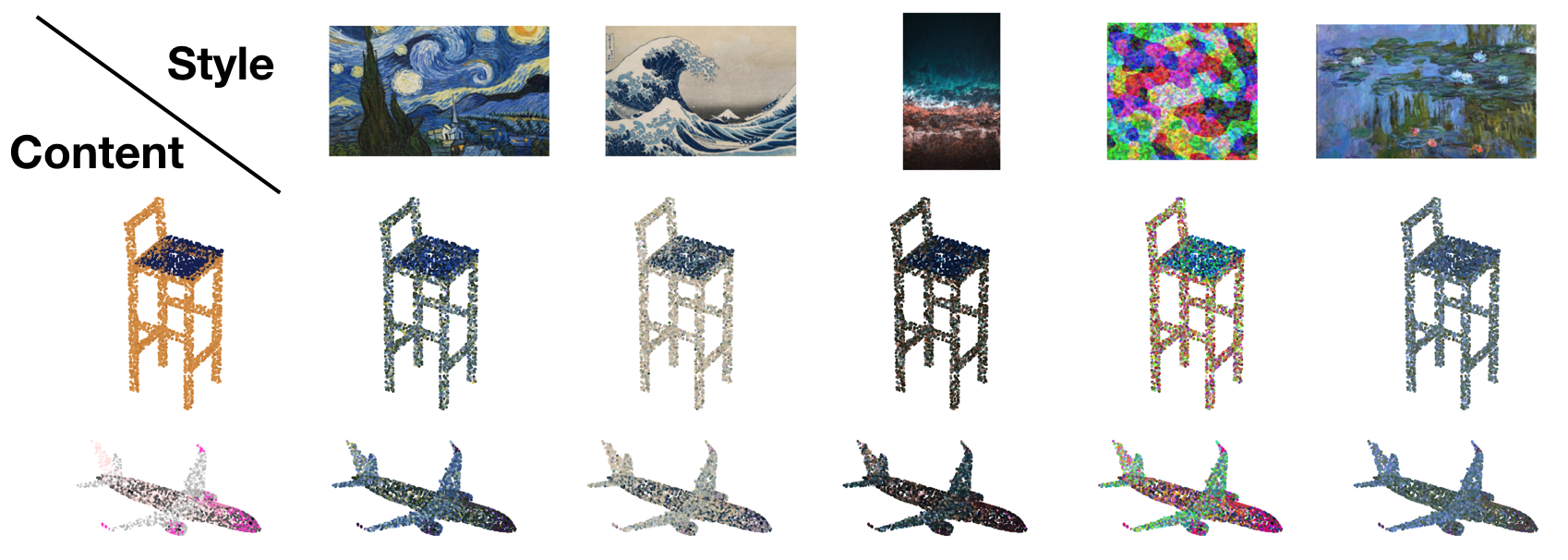}\label{object}}\\
\subfloat[][]{\includegraphics[width=\linewidth]{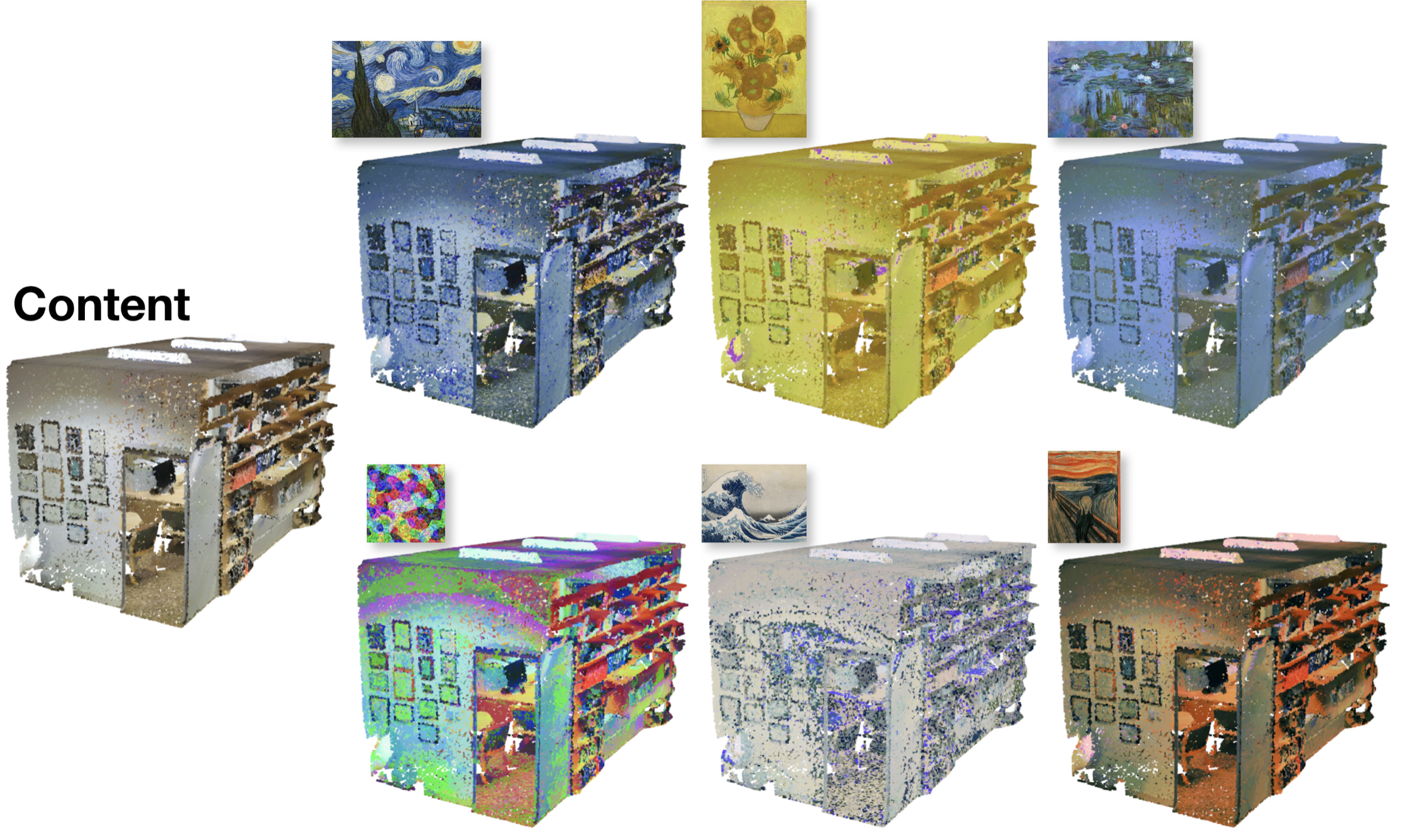}\label{scene}}
\end{center}
\caption{Style transfer results from images onto point clouds. \protect\subref{object} Single object point clouds with 4096 points. \protect\subref{scene}}Indoor scene point clouds with 160k points. Both results demonstrate that the color pattern of point clouds (e.g., darker seat of the chair) are preserved while the color distribution is more like that in the style image.
\label{fig:img_transfer}
\end{figure}

In this experiment, we stylise the color expression of content point clouds $C_{color}$ either of single objects or indoor scenes from images. The style image is reshaped to $(H*W)  \times 3$ and being treated as $S_{color}$. In both cases, content representation is extracted from \{FEL-64\} and style representations are extracted from \{FEL-1024, FEL-2048\}. $\alpha_{color}$ is 1, and $\beta_{color}$ is 100. $P_{color}$ is initialized as $C_{color}$. Qualitative results on point clouds of single objects are given in Fig.\ref{fig:img_transfer}(a).
The point cloud of indoor scenes is from S3DIS dataset \cite{S3DIS_cvpr16} and down-sampled to 160k points for the sake of computation budget. Adam optimizer with learning rate 0.001 is utilized to update $P_{color}$ for 30000 steps. Quantitative results are given in Fig.  \ref{fig:img_transfer}(b).
\subsubsection{Style transfer loss}
\begin{figure}[htb]
\begin{center}
\subfloat[][]{\includegraphics[width=0.5\linewidth]{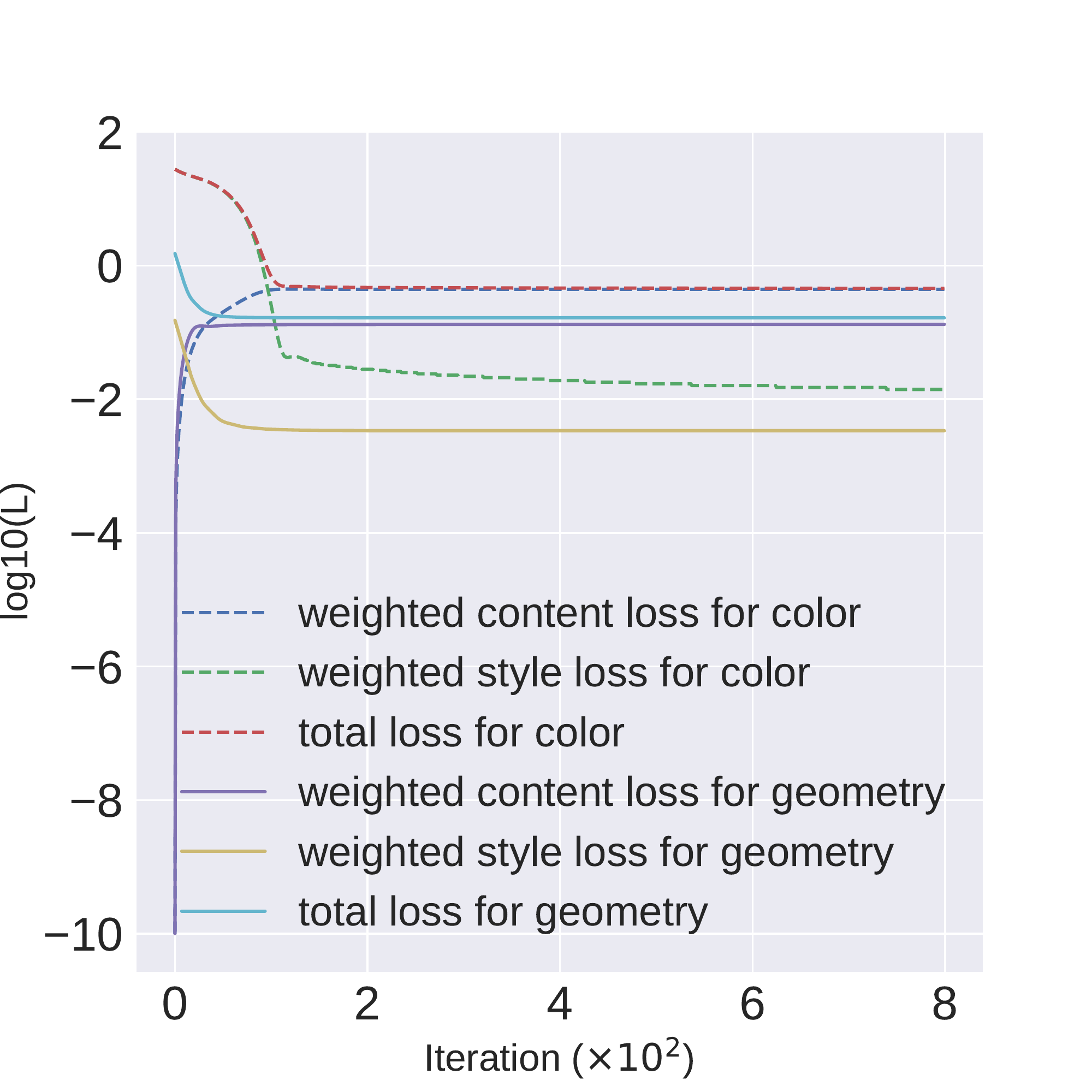}\label{object}} 
\subfloat[][]{\includegraphics[width=0.5\linewidth]{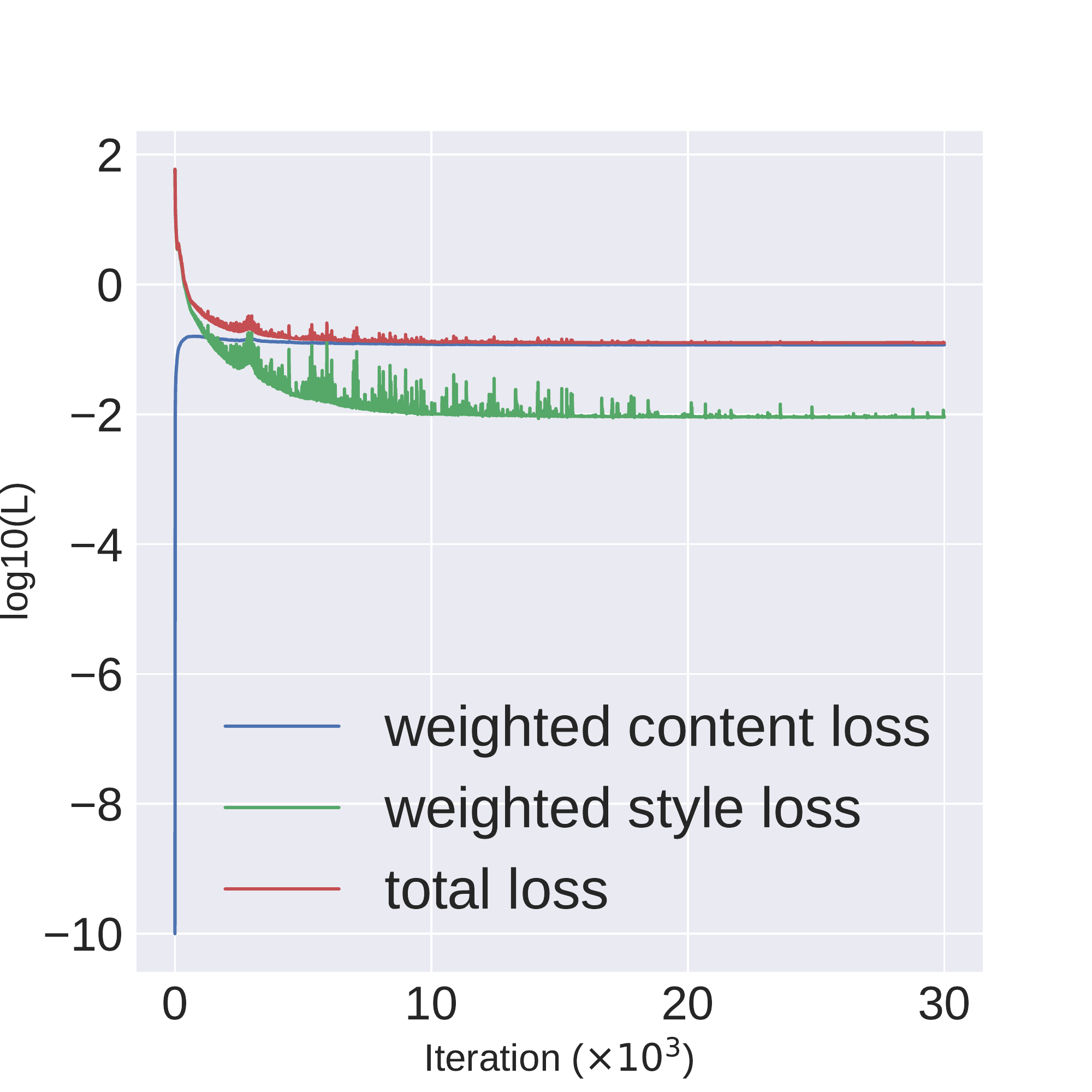}\label{scene}}
\end{center}
\caption{Change of style transfer loss.
\protect\subref{object} Style transfer on the geometry and color property between single objects.
\protect\subref{scene} Style transfer on the color property of indoor scene point clouds. In both cases, the total losses converge fast to stability.}
\label{fig:style_transfer_loss}
\end{figure}

To confirm whether the iterative update process eventually converge, we visualize the style transfer loss in Fig.\ref{fig:style_transfer_loss}. Fig.\ref{fig:style_transfer_loss}(a) illustrates an example of the change in the loss during the stylisation between single object point clouds in Fig.\ref{fig:pc_transfer}(a). We can find that the style transfer loss converge fast to stability within just 200 iterations. Since $P$ is initialized as $C$, the content loss is 0 at the beginning and increases to stability.

The loss of transferring from images onto color expression of point clouds is given in Fig.\ref{fig:style_transfer_loss}(b). In this case, the fact that indoor scene point clouds comprise much more points detains the update process. This difficulty is the reason why we decrease the initial learning rate to 0.001 and increase the iteration steps to 30000 in Sec.\ref{sec:color_transfer_from_images}.
\subsection{Analysis of style transfer }
\label{sec:analysis_of_style_transfer}
According to Eq.\ref{eq:geo_total} or Eq.\ref{eq:color_total}, given a pre-trained network for feature extraction, four factors are likely to affect the stylised point cloud: the weight of content loss and style loss, the layer used to extract feature representation, the initialization strategy of $P$ and the choice of the optimizer to seek $P_{geo}^{\mbox{*}}$ or $P_{color}^{\mbox{*}}$. In this section, we investigate their effects on the stylised point clouds.
\subsubsection{Content-style trade-off}
\begin{figure}[tb]
\begin{center}
\subfloat[][]{\includegraphics[width=\linewidth]{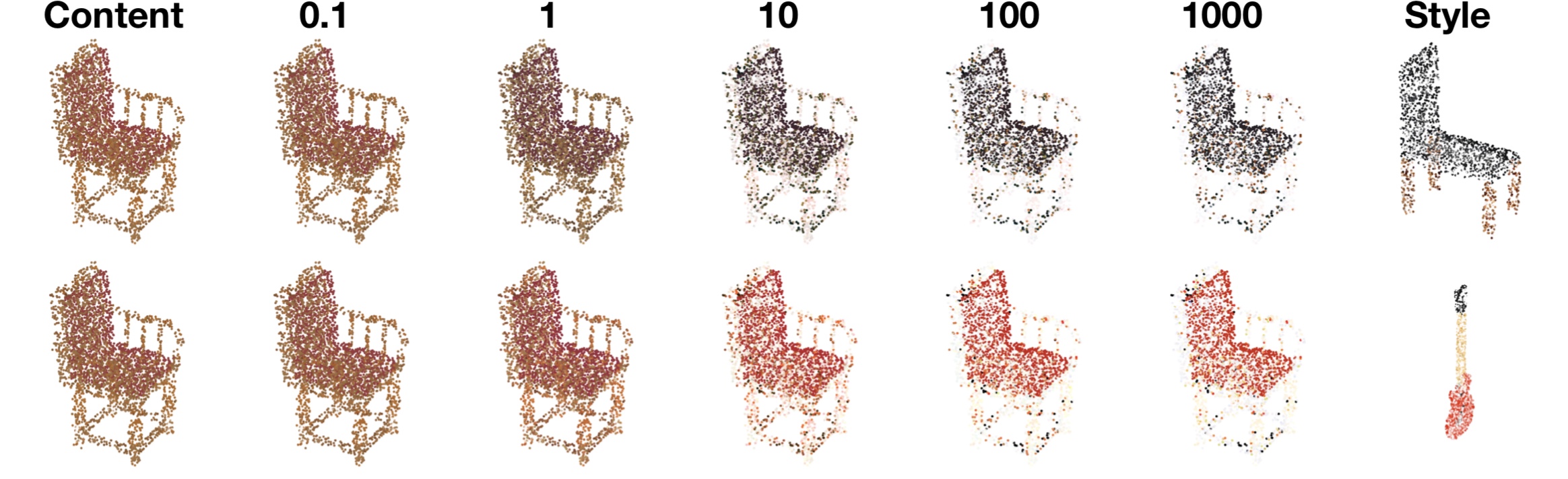}\label{ratio_color}}\\
\subfloat[][]{\includegraphics[width=\linewidth]{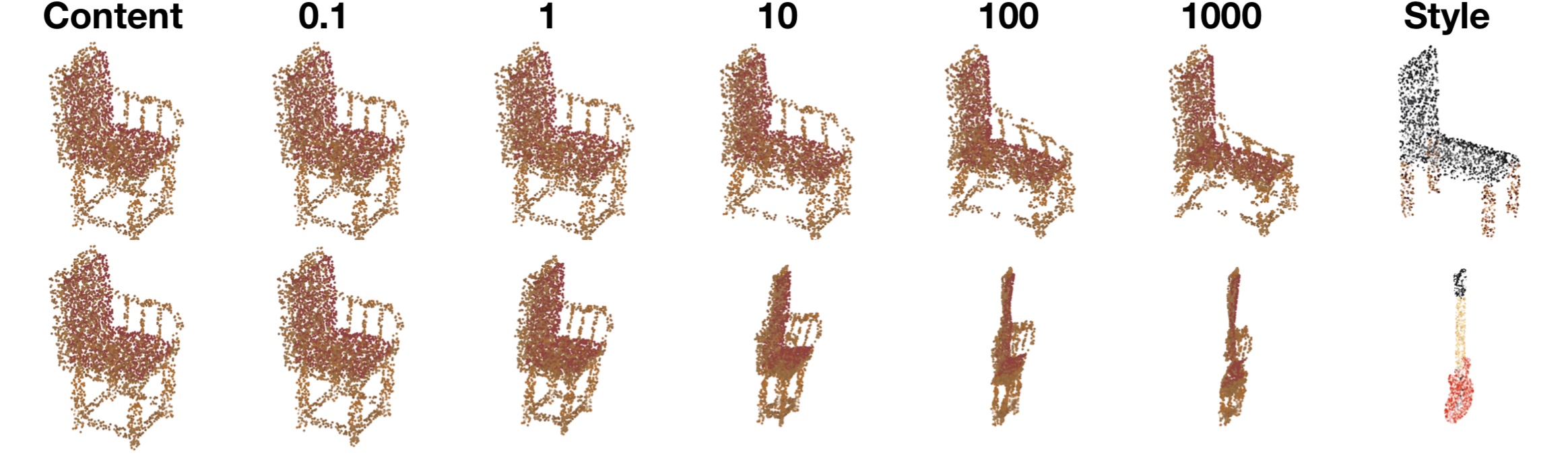}\label{ratio_geo}}
\end{center}
\caption{Style transfer results with different $\beta$. 
\protect\subref{ratio_color} Transfer on color property with different $\beta_{color}$.
\protect\subref{ratio_geo} Transfer on geometry property with different $\beta_{geo}$.}
\label{fig:ratio}
\end{figure}

Intuitively, different ratios between the content loss and style loss may decide how much the stylised point cloud look like the content or the style point cloud. We extract the content representation and style representation both from \{FEL-64\}; $P$ is initialized as $C$. In the first experiment, we fix $P_{geo}$ as $C_{geo}$, and $P_{color}$ is to be stylised. We fix $\alpha_{color}$ as 1 and vary $\beta_{color}$ among 0.1, 1, 10, 100 and 1000. The results are shown in Fig.\ref{fig:ratio} (a). 

In the second experiment, we fix $P_{color}$ as $C_{color}$, and $P_{geo}$ is to be stylised. $\alpha_{geo}$ is set to 1 and $\beta_{geo}$ is varied among 0.1, 1, 10, 100 and 1000. The results are given in Fig.\ref{fig:ratio} (b). From the results in both experiments, we can see that as $\beta$ increases, the style of $S$ is more evidently transferred onto $P$ while its content is preserved. In the case of color property, the color pattern of the content chair point cloud (darker seat) is preserved in all cases while its overall color distributions become more similar to that of $S_{color}$. In the case of geometry property, the topology of the content chair point cloud is preserved while its overall shape becomes more similar to the style point cloud.
\begin{figure}[htb]
\begin{center}
\subfloat[][]{\includegraphics[width=\linewidth]{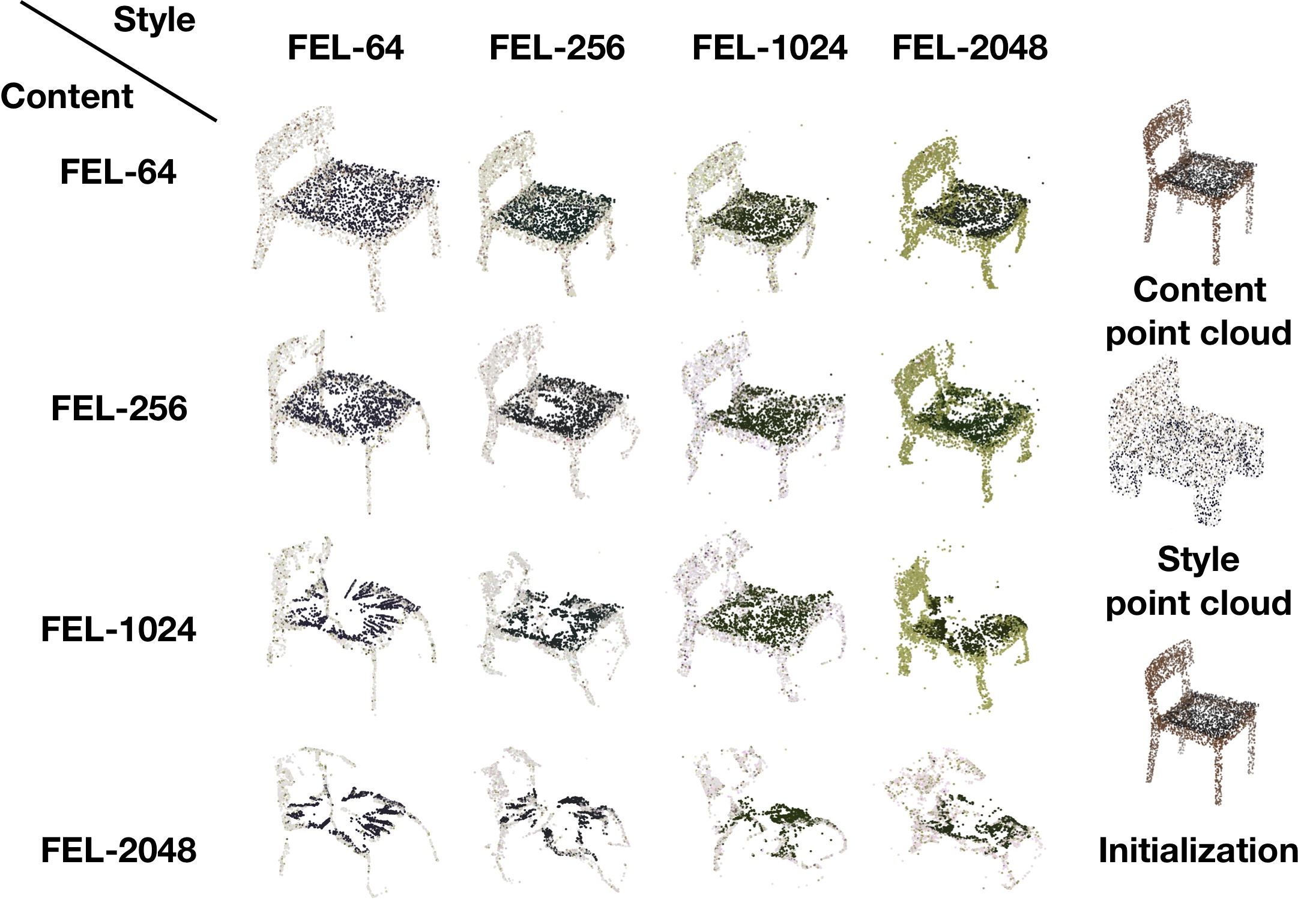}\label{content}}\\
\subfloat[][]{\includegraphics[width=\linewidth]{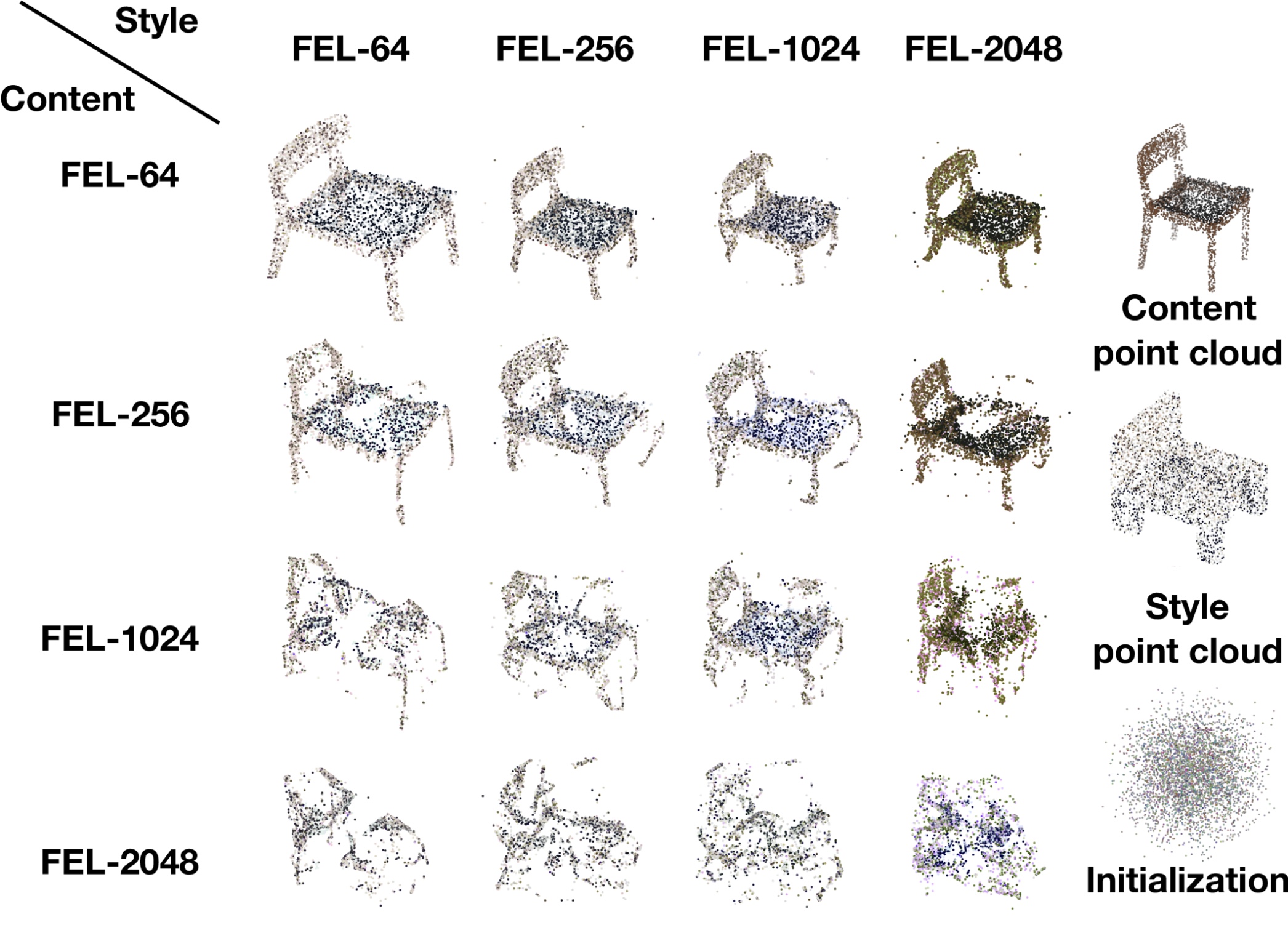}\label{random}}
\end{center}
\caption{Style transfer results with content/style representation extracted from different layers and different initialization strategy.
\protect\subref{content} Initialize $P$ as content point cloud.
\protect\subref{random} Randomly initialize $P$. Initializing $P$ as $C$ helps maintaining the shape.}
\label{fig:layer_test}
\end{figure}
\subsubsection{Content/style representation from different layers}
\label{sec:layer_test}
In this experiment, we aim at inspecting the effect of target feature extracted from different layers on the style transfer results. Explicitly, we compute style loss and content loss based on features extracted from only one FEL but enumerate all their combinations. 
There are 4 FELs from which we can extract content or style representation, resulting in 16 combinations. We set $\beta_{color} $ and $\beta_{geo}$ as 100 and 10, respectively. The results are shown in Fig.\ref{fig:layer_test} (a). The results in a row are computed from the same content representation while those in the same column are computed from the same style representation; the column or the row name denotes the layer name. 
As we can see, when content representations are extracted from low layers, the generated point cloud preserve a distinct shape; when content representations extracted from high layers, the overall shape of generated point clouds become vague. 

In the case of style representation, when they are extracted from low layers, the point cloud is more stylised; when they are extracted from high layers, the stylised point clouds again become more similar to the content point cloud, and noise-like points appear.
\subsubsection{Initialization strategy }
As the stylisation is an iterative update process on $P$, different initialization strategy is likely to lead to different $P^\star$. So far we have initialized $P$ as $C$, we compare the initialization strategy with an alternative initialization method: every element of $P$ is sampled from $\mathcal{N}(0, 0.5)$. We conduct the same experiment as in \ref{sec:layer_test} except for the different initialization strategy, as shown in Fig.\ref{fig:layer_test}(b). 

Compare to Fig.\ref{fig:layer_test}(a), we find that when feature representations are extracted from low layers, different initialization strategies result in almost the same stylised point cloud, however, initializing $P$ as $C$ helps $P$ maintain a distinct shape. When $P$ is randomly initialized, and feature representations are extracted from high layers, the result becomes blurred and unrecognizable, but its counterpart in Fig.\ref{fig:layer_test}(a) still have a relatively clear shape.
\subsubsection{Choice of optimizer}
\begin{figure}[tb]
\begin{center}
\subfloat[][]{\includegraphics[width=0.5\linewidth]{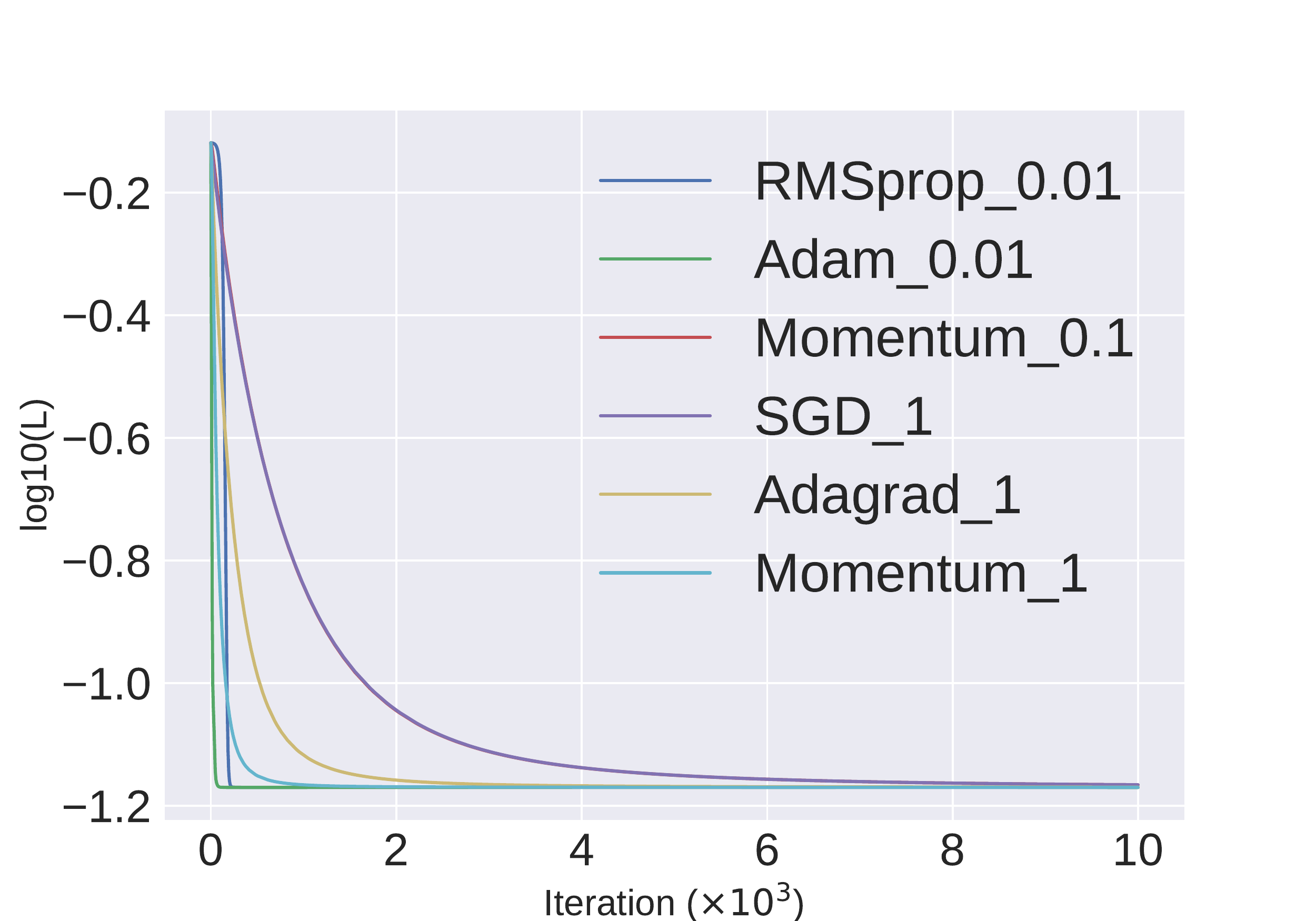}\label{geo}} 
\subfloat[][]{\includegraphics[width=0.5\linewidth]{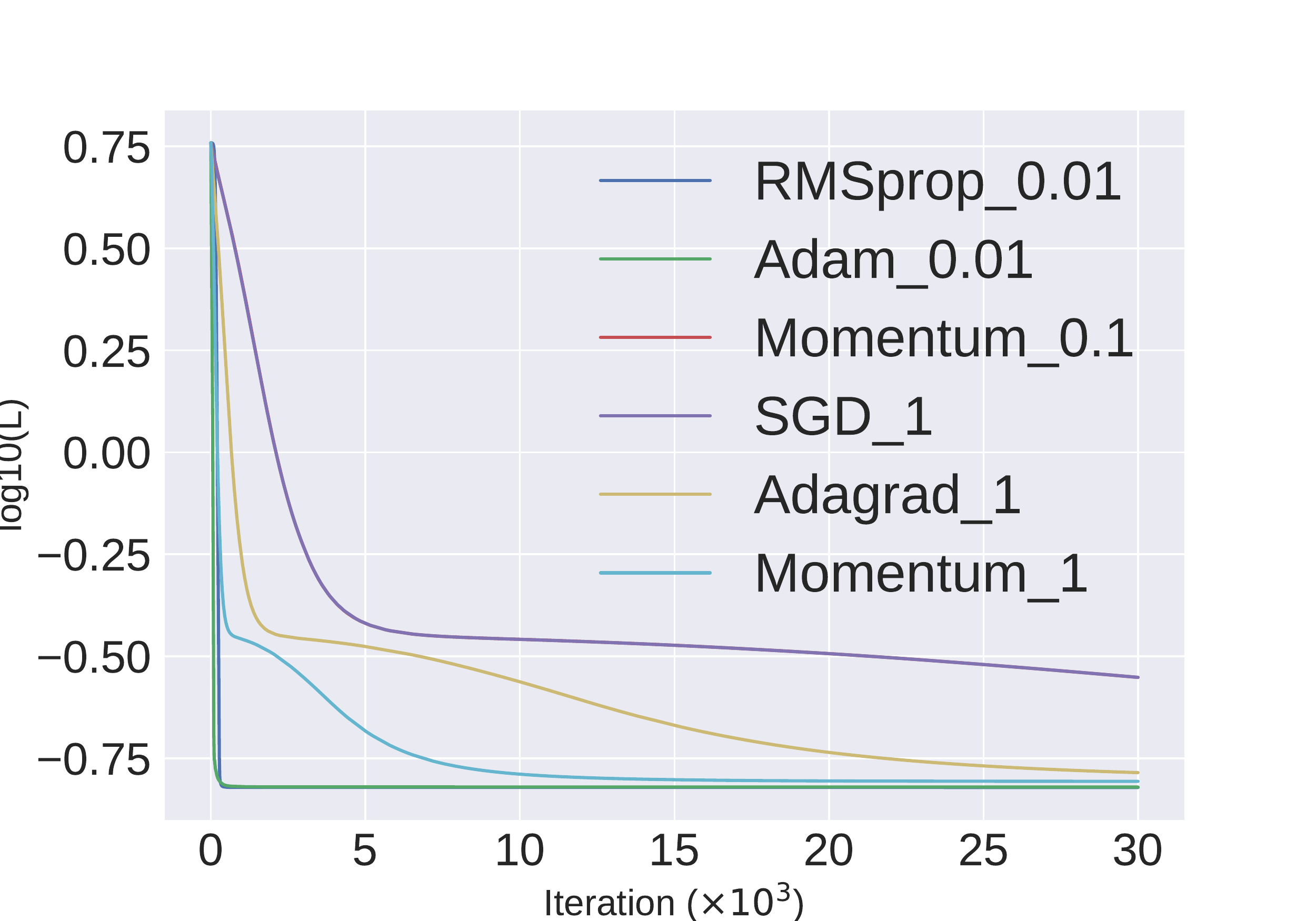}\label{color}} \\
\subfloat[][]{\includegraphics[width=\linewidth]{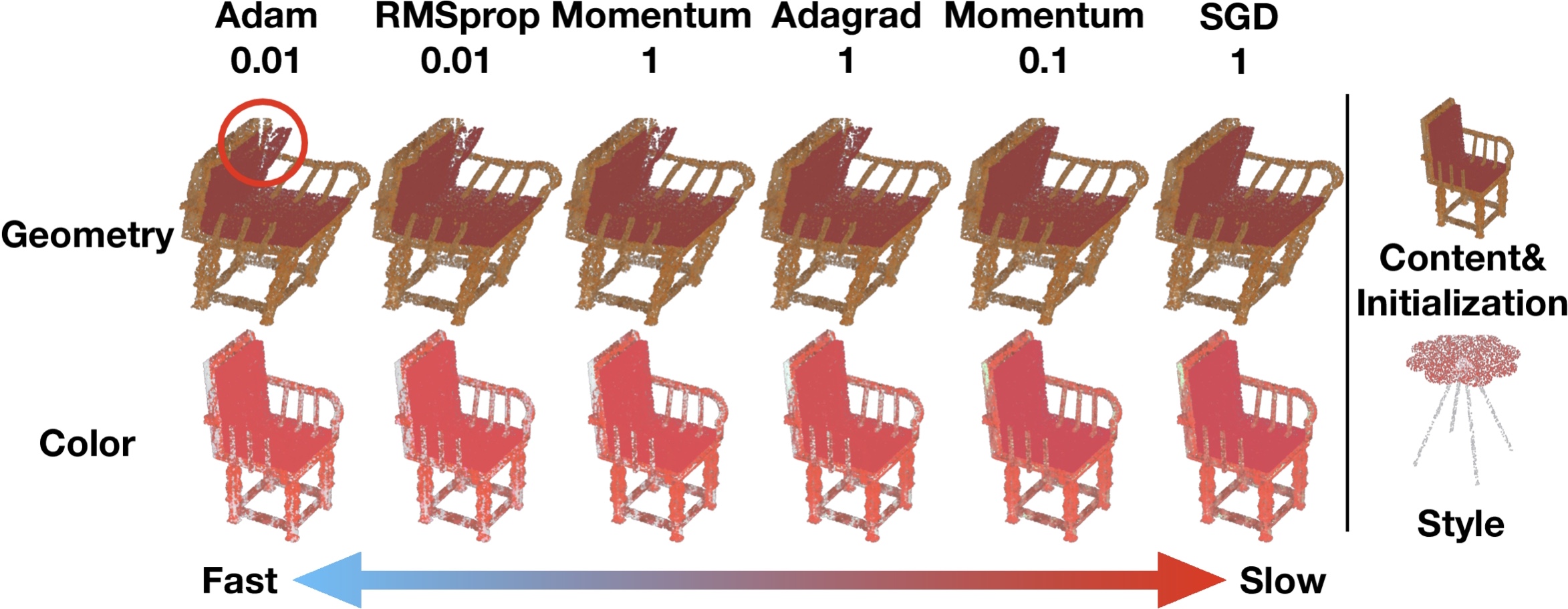}\label{optimizer}}
\end{center}
\caption{Stylise point clouds by different optimizers.
\protect\subref{geo} Style loss curve of geometry.
\protect\subref{color} Style loss curve of color.
\protect\subref{optimizer} Qualitative comparison of stylised point clouds updated by different optimizers. There is no apparent difference in overall shape or color distribution. However, cracks appear in point clouds updated by fast-converged optimizers.
}
\label{fig:optimizer_style_transfer_loss}
\end{figure}

Different optimizers with different learning rate affect the time for $P$ to convergence and may result in different stylised point clouds. In this experiment, we utilized different optimizers from the combination of the optimizer in \{SGD,   Momentum \cite{qian1999momentum}, Adagrad \cite{adagrad}, RMSprop, Adam \cite{Kingma2014AdamAM}\} and the learning rate in \{0.01, 0.1, 1\} to update $P$.  We filter out those not converged after 30k iterations either because of slow learning rate or fluctuation and illustrate the style transfer loss curve of the remaining in Fig.\ref{fig:optimizer_style_transfer_loss} (a) and (b). It is evident that whatever the speed to converge, the final content loss of geometry is almost the same. 

We give a qualitative comparison of the stylised point cloud by different optimizers in Fig.\ref{fig:optimizer_style_transfer_loss}(c). We find that fast convergence does not mean a better quality of the stylised point clouds. As shown in the red circle of Fig.\ref{fig:optimizer_style_transfer_loss}(c), cracks appear in the point clouds updated by fast-converged optimizers, e.g., Adam. The appearance of cracks make the point cloud look like being chopped up. In contrast, the problem is alleviated or disappears in the point clouds updated by slow-converged optimizers, e.g., SGD. We consider the reason of the appearance of these cracks is likely to be that fast update encourages the optimizer to merely translate different parts of a point cloud along different directions to minimize the style loss. 
\subsection{Ablation study on model architecture}
\label{sec:ablation_study}
As described in Sec.\ref{sec:classification_network}, we add two modifications to PointNet: the separation of different kind of properties of point clouds (late-fusion) and the replacement of shared fully connected layer with FEL. In this section, we evaluate the effectiveness of our modifications for point cloud style transfer.
\subsubsection{Comparison on classification performance}
Although the primary goal of modifying PointNet is to extract better representations for style transfer, we are still interested in whether the two modifications improve the model performance on colored point clouds classification task. We compare our final model with two alternative models: the model directly consuming colored point cloud without splitting it (early-fusion) and the model without using FEL (the same as PointNet except for late-fusion). We report the performance of these models in Table.\ref{tab:acc} based on two evaluation metrics: accuracy and multiclass Area Under the ROC Curve (AUC). According to Table.\ref{tab:acc}, our final model achieves the best performance, which validates our design.
\begin{table}[tb]
\centering
\begin{tabular}{lcc}
\hline
 model type  &accuracy  &multiclass AUC  \\
 \hline
 early-fusion + FEL  &90.54  &98.49  \\
 late-fusion  &87.50  &96.56  \\
 \hline
 late-fusion + FEL  &\textbf{90.80}  &\textbf{98.85} \\
 \hline
\end{tabular}
\caption{Classification performance of different model choices on DensePoint test set \protect\cite{densepoint}.}
 \label{tab:acc}
\end{table}

\begin{figure}[tb]
\begin{center}
\includegraphics[width=\linewidth]{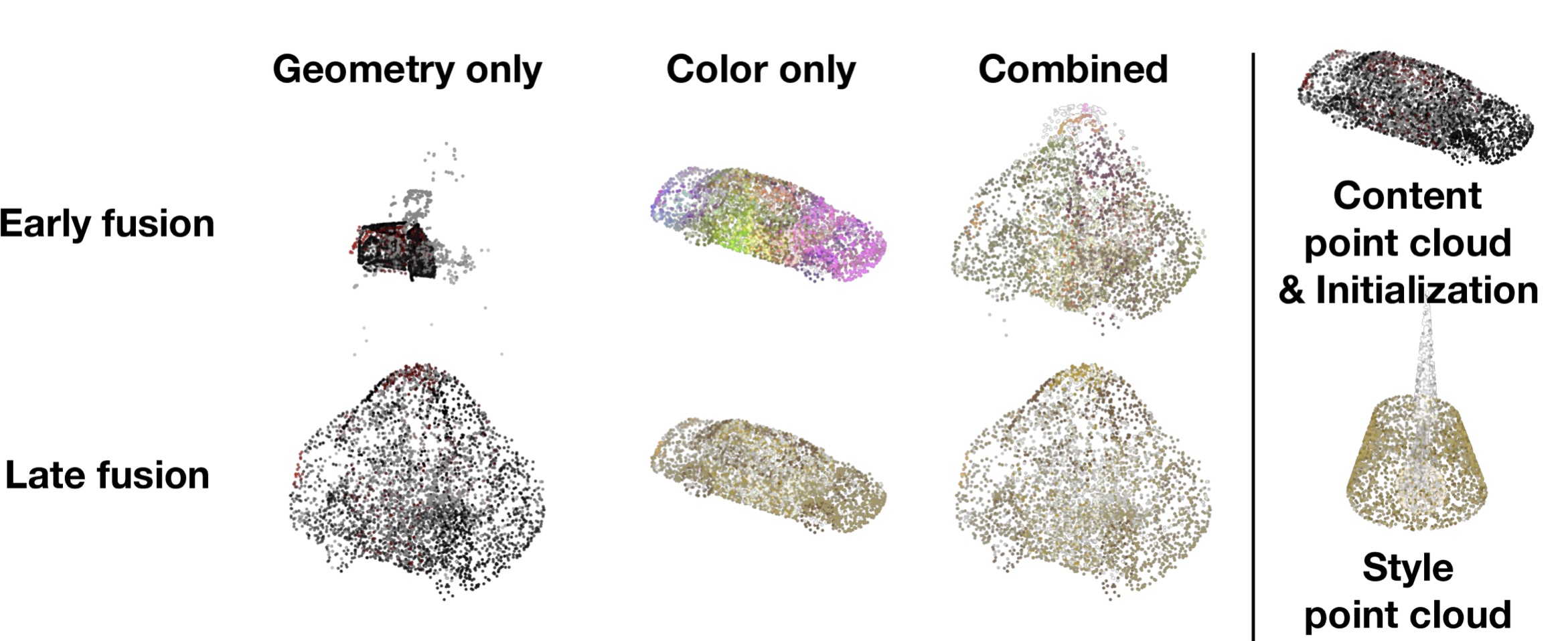}
\end{center}
\caption{Style transfer results comparison on different model architecture designs. }
\label{fig:earlyF_lateF}
\end{figure}
\subsubsection{Late fusion v.s. early fusion }
We design a two-route network to independently process the geometry and color property of point clouds and concatenate extracted two global feature vectors later (late fusion). An alternative network design is a one-route network directly consuming colored point clouds which are not split into two parts (early fusion). In the latter case, we cannot extract feature representations separately for geometry or color property;Eq.\ref{eq:geo_content} and Eq.\ref{eq:color_content} degrade to: 
\begin{equation}
L_{content}(P, C)=\sum_{l\in \{l_c\}}{||F^l(P)-F^l(C)||^2} \label{eq:earlyF_content}
\end{equation}
, and Eq.\ref{eq:geo_style} and Eq.\ref{eq:color_style} degrade to
\begin{equation}
L_{style}(P, S)=\sum_{l\in \{l_s\}}{||G(F^l(P))-G(F^l(S))||^2} \label{eq:earlyF_style}
\end{equation}However, we can still opt to update only the geometry or color property and fix another part during the update process as follows:
\begin{equation}
\begin{split}
\newcommand{\argmin}{\mathop{\rm arg~min}\limits}
P_{geo}^{\mbox{*}}=\argmin_{P_{geo}} \alpha L_{content} (P, C) + \beta L_{style}(P, S)  \label{eq:earlyF_total}
\end{split} 
\end{equation}
, where $P_{geo}$ can be replaced by $P_{color}$ or $P$ depending on whether we want to modify the geometry property, the color property or both of the point cloud $P$.

We conduct a comparison experiment between the two models to inspect their effects on the style transfer results, as shown in Fig.\ref{fig:earlyF_lateF}. In both cases, content/style representations are extracted from the first layer, and $P$ is initialized as $C$. In the case of early fusion network, $\alpha = 1, \beta = 100$;in the case of late fusion network, $\alpha_{color} = \alpha_{geo} = 1, \beta_{geo} = 100, \beta_{color} = 1000$.

As we can see in Fig.\ref{fig:earlyF_lateF}, in the case of early fusion network, only updating the geometry or the color property leads to absurd results. This is not surprising since the extracted representation depends on both the geometry and color property of $P$. When the geometry and the color property are both updated, the result is somehow close to that from late fusion networks, but still not appealing. In contrast, by utilizing a late fusion network, either the geometry or the color property can be modified, resulting in appealing stylised point clouds. The comparison demonstrates the effectiveness and flexibility of our proposed late fusion design. 
\subsubsection{FEL v.s. shared FC}
\begin{figure}[htb]
\begin{center}
\includegraphics[width=\linewidth]{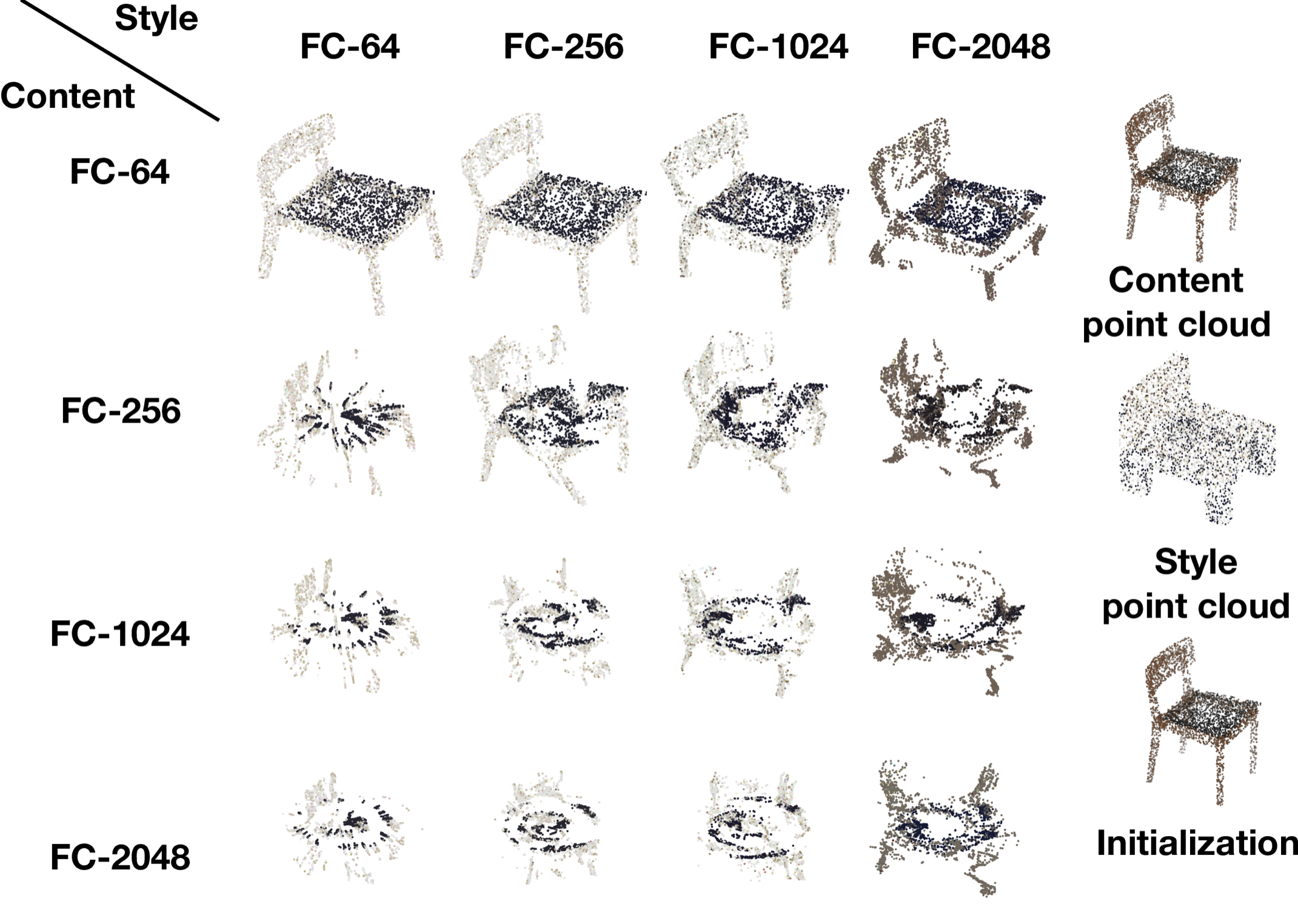}
\end{center}
\caption{Style transfer results by using the model without FEL. }
\label{fig:wo_FEL}
\end{figure}

When the point-wise concatenation operation in Fig.\ref{fig:model}(b) is not performed, a feature encoding layer degrades to a naive shared fully connected (shared FC) layer as in PointNet. We can still extract feature representations in this design and perform style transfer in the same way as described in Sec.\ref{sec:transfer_between_point}. The same experiment as in Sec.\ref{sec:layer_test} is conducted, as shown in Fig.\ref{fig:wo_FEL}. We can find that when content representations are extracted from the first layer, the quality of stylised point clouds is comparable to that in Fig.\ref{fig:layer_test}(a). However, when the content representation is extracted from higher layers, the quality dropped drastically wherever the style representation is from. This comparison validates the robustness of our design of replacing shared FCs in a naive PointNet with FELs.

\section{Discussion and Future Work}
\label{sec:discussion}
In this study, we propose a neural style transfer method for point clouds that can independently stylise its geometry and color property. The method is extended to stylise the color property of a point cloud from the style of an image. Experimental results show that the method works well even when the number of points in the content point cloud and the style point clouds/image is different. 

We think our method is intrinsically to perform distribution matching between content point clouds and style point clouds either on geometry or on color property. It is evident in the case of color transfer that the color of a content point cloud is likely to be replaced based on their empirical distribution correspondence. However, this overall distribution matching introduces a problem that the transfer ignores semantic correspondence when both point clouds comprise the same functional parts, e.g., to stylise a seat of a chair from a seat of another chair. To address this problem is one of the routes in future work.

Another route is to accelerate the stylisation process. The generation of a stylised point cloud requires a time-consuming iterative update process. We can take insight from the evolution of image style transfer methods that an additional network can be trained to generate a stylised point cloud with only one feed-forward pass.

{\small
\bibliographystyle{ieee}
\bibliography{refs}
}
\end{document}